\documentclass[letterpaper, 10 pt, conference]{ieeeconf}  % Comment this line out if you need a4paper

\IEEEoverridecommandlockouts                              % This command is only needed if 
\usepackage{cite}
\usepackage{multicol}
\usepackage[bookmarks=true]{hyperref}
\usepackage{hyperref}
\usepackage{graphics} % for pdf, bitmapped graphics files
\usepackage{times} % assumes new font selection scheme installed
\usepackage{amsmath} % assumes amsmath package installed
\usepackage{amssymb}  % assumes amsmath package installed
\usepackage{subfigure}
\usepackage{caption}
\usepackage{pgf, tikz, pgfplots}
\usepackage{tabularx,longtable,multirow}%hangcaption
\usepackage{diagbox}
\usepackage{pifont}

\newcommand{\DTWOCOPLAN}{\texttt{\textsc{D2CoPlan}}}
\newcommand{\EXPERT}{\texttt{\textsc{Expert}}}
\newcommand{\DG}{\texttt{\textsc{DG}}}
\newcommand{\DMP}{\texttt{\textsc{DMP}}}
\newcommand{\ORACLE}{\texttt{\textsc{Oracle}}}

\title{\LARGE \bf
D2CoPlan: A Differentiable Decentralized Planner\\ for Multi-Robot Coverage
}

\author{Vishnu Dutt Sharma$^{1}$, Lifeng Zhou$^{2}$, and Pratap Tokekar$^{1}$% <-this % stops a space
\thanks{This work is supported by the National Science Foundation under Grant No. 1943368 and ONR under grant number N00014-18-1-2829.}% <-this % stops a space
\thanks{$^{1}$Vishnu D. Sharma and Pratap Tokekar are with the Dept. of Computer Science,
        University of Maryland, College Park, MD, USA
        {\tt\small \{vishnuds, tokekar\}@umd.edu}}%
\thanks{$^{2}$Lifeng Zhou is with the  Dept. of Electrical and Computer Engineering, Drexel University, Philadelphia, PA, USA
        {\tt\small lz457@ drexel.edu}}%
}

\begin{document}

\maketitle
\thispagestyle{empty}
\pagestyle{empty}

%%%%%%%%%%%%%%%%%%%%%%%%%%%%%%%%%%%%%%%%%%%%%%%%%%%%%%%%%%%%%%%%%%%%%%%%%%%%%%%%
\begin{abstract}
Centralized approaches for multi-robot coverage planning problems suffer from the lack of scalability. Learning-based distributed algorithms provide a scalable avenue in addition to bringing data-oriented feature generation capabilities to the table, allowing integration with other learning-based approaches. To this end, we present a learning-based, differentiable distributed coverage planner (\DTWOCOPLAN) which scales efficiently in runtime and number of agents compared to the expert algorithm, and performs on par with the classical distributed algorithm. In addition, we show that \DTWOCOPLAN{} can be seamlessly combined with other learning methods to learn end-to-end, resulting in a better solution than the individually trained modules, opening doors to further research for tasks that remain elusive with classical methods.
\end{abstract}

%%%%%%%%%%%%%%%%%%%%%%%%%%%%%%%%%%%%%%%%%%%%%%%%%%%%%%%%%%%%%%%%%%%%%%%%%%%%%%%%
\section{Introduction}
Multi-robot coverage and tracking is a well-studied problem. Over the years, several approaches have been presented for planning and coordination algorithms~\cite{zhou2022multi}. In particular, consider the problem of covering a set of mobile targets using a team of aerial robots with downwards-facing cameras (Figure~\ref{fig:uav_tracking}). A target is said to be covered if it falls within the field-of-view of one of the robots' cameras. The objective is for the robots to choose their individual trajectories so as to maximize the total number of targets covered. 

There are several reasons why this problem is challenging. Coordination amongst the robots is critical as you want to avoid overlap and maximize the coverage. This is easier in a centralized setting; however, our focus is on \emph{decentralized} strategies where the robots can communicate directly only with their immediate neighbors. Decentralization is also harder since each robot only knows of the targets in their own fields-of-view. Finally, since we need to cover mobile targets, we need to predict their motion over the planning horizon. However, the motion model of the targets itself may be unknown making the problem even more challenging.

In this paper, we investigate the question: \emph{Can the robots \emph{learn} to plan and coordinate in a decentralized fashion for target coverage problems?} Recently, there has been significant work on learning-based approaches to multi-robot planning. However, most of this work is restricted to coordination for path finding (where each robot needs to find the shortest path to its own goal position in an unknown environment)~\cite{gama2020graphs,li2020graph,li2021message} and formation control (such as flocking)~\cite{Tolstaya19-Flocking,khan2019graph}. We build on this to study a more complex task that requires planning, coordination, and prediction.

Our contribution is a decentralized, differentiable coverage planner (\DTWOCOPLAN{}) for multi-robot teams. \DTWOCOPLAN{} consists of three differentiable modules, namely map encoder, decentralized information aggregator, and local action selector. The input to \DTWOCOPLAN{} is a coverage map that represents predictions of where the targets are going to be in the next timestep. This map comes from another differentiable module we call Differentiable Map Predictor (\DMP{}). The map encoder takes the predicted maps and turns it into a compact representation which is shared with the other agents using a Graph Neural Network (GNN)~\cite{scarselli2008graph}. The GNN aggregates information from neighboring agents and uses that for selecting the ego robot's action. \DTWOCOPLAN{} is trained on an expert strategy (centralized optimal algorithm that has global information) but is executed in a decentralized fashion (following the Centralized Training, Decentralized Execution paradigm~\cite{kraemer2016multi}). We show that \DTWOCOPLAN{} is a scalable, efficient approach for multi-robot target coverage. In particular, we show that \DTWOCOPLAN{} is able to achieve $~93\%$ of the centralized optimal algorithm in upto $~150$x less time but in a decentralized fashion. 

A typical approach for this problem is to frame it as a submodular maximization problem with a uniform matroid constraint~\cite{calinescu2011maximizing}. A decentralized greedy (\DG{}) algorithm gives theoretical performance guarantees and works well empirically~\cite{qu2019distributed,williams2017decentralized}. We show that \DTWOCOPLAN{} performs as well as \DG{} when the ground truth positions of the targets are known and better when the robots have to predict the motion of the targets. Further, the running time of \DTWOCOPLAN{} scales better compared to \DG{}. A key advantage of \DTWOCOPLAN{} is that it consists of two differentiable modules, where the observation processor module can be trained to be compatible with the planner module. We investigate several ways of combining the two modules as well as ablation studies for the design of \DTWOCOPLAN{}'s architecture.

The rest of the paper is organized as follows:
we first discuss the related work on this topic in Section~\ref{sec:related_work}. Then we formulate the problem in Section~\ref{sec:problem_formulation}. Section~\ref{sec:methods} describes the design of \DTWOCOPLAN{}. Section~\ref{sec:experiment_results} first provides the implementation details and then describe various experiment and the results obtained. We conclude by summarizing our finding in Section~\ref{ref:conclusion} and discuss the avenues of future work.

\section{Related Work}\label{sec:related_work}
Multi-robot coordination problems have largely relied on using classical, non-leaning-based approaches. The centralized approaches assume presence of a single entity which can access observations from all the robots and plan accordingly. Since finding optimal solutions may be practically intractable, the centralized approaches often utilize greedy formulations to find approximate solutions. Many multi-robot tracking and coverage objectives are submodular i.e., they have diminishing return property, and greedy solutions provide constant factor approximation guarantee for them~\cite{zhou2021multi}. 

Finding solutions with centralized approaches is still computationally expensive and the runtime rapidly increases with the increase in the number of robots. Decentralized approached provide an efficient solution at the cost of a lower, but acceptable drop in the task performance, by distributing the task of computation to cliques\cite{zhou2021multi,zhou2022distributed,shi2021communication}. The communication could be expanded to multiple hops to increase the information horizon, but it comes at the cost of increased runtime~\cite{qu2019distributed,williams2017decentralized}.

Neural networks provide an avenue to improve upon classical solutions through their ability to model complexities using data. Furthermore, a differentiable approach can be combined with other differentiable methods to enable efficient with end-to-end learning from data~\cite{chaplot2021differentiable}. Introduction of GNNs~\cite{scarselli2008graph} to solve problem with graph representations opened doors to application of neural networks to decentralized multi-robot tasks by facilitating feature sharing between robots~\cite{gama2020graphs}. Recent works have successfully employed GNNs to solve multi-robot problems such path planning~\cite{li2020graph, li2021message}, persistent monitoring~\cite{chen2021multi}, and formation control~\cite{Tolstaya19-Flocking, khan2019graph} among others. Specifically for multi-robot coverage problems, Tolstaya et al.~\cite{tolstaya2021multi} and Gosrich et al.~\cite{gosrich2022coverage} used GNNs in different training paradigms to learn control policies. Many of these works show that apart from achieving near-expert solutions, GNNs can help scale well to larger robot teams.

Unlike these works, we specifically focus on target coverage. Recently, Zhou et al.~\cite{zhou2021graph} proposed a planner for the coverage problem using GNN and show such a planner performs on par with the classical counterpart and scales marginally better. However, their approach requires hand-crafted features and uses only 20 closest target as input. This design makes the network non-differentiable at the input layer and thus can not be used in conjunction with other learning methods. We address both these issues in our work by using a richer map representation, while also improving the coverage performance and scalability. 

\section{Problem Formulation}\label{sec:problem_formulation}
\begin{figure}
\centering
\includegraphics[width=0.65\columnwidth]{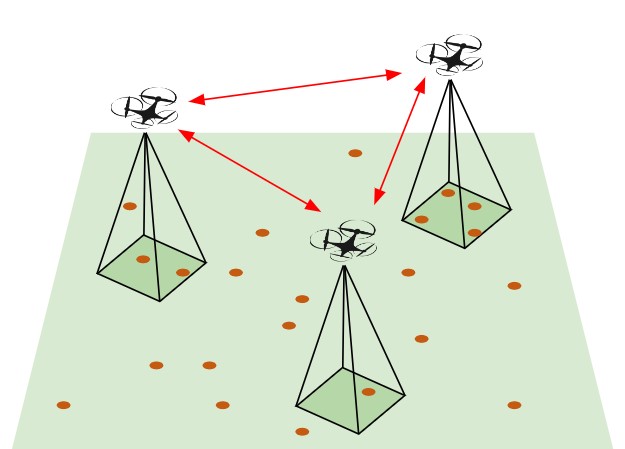}
\caption{\small Multi-robot target coverage: a team of aerial robots aims at covering multiple targets (depicted as red dots) on the ground.
The robots observe the targets in their respective field of view  (green squares) using down-facing cameras and share information with the neighbors by communication links (red arrows).
\label{fig:uav_tracking}}
\end{figure}

In this work, we investigate the problem of decentralized, multi-robot action selection for joint coverage maximization. Consider the scenario in Figure~\ref{fig:uav_tracking}. 
% A team of aerial robots is tasked with covering multiple mobile targets on the ground. Each robot observes a limited part of ground through down-facing cameras mounted on it, and can communicate with other robots within its communication range. There isn't any central entity to facilitate planning and thus the robots must collaborate with each other to execute actions that maximize the joint coverage by all the robots. 
A set of $N$ robots are tasked to cover targets moving in a grid of size $G\times G$. Every robot $R_i$ has a set of actions $A_i$ that it must select from at each time step. All the targets that fall within the sensing range $r_s$ (e.g., camera footprint) are said to be \textit{covered} by the robot. The objective is to maximize the total number of targets covered by selecting the actions for each robot. 

We assume that the robots do not collide with each other (e.g., by flying at different altitudes). A robot $R_i$ can communicate with another robot $R_j$ is if it is within the \textit{communication range} $r_c$. 
% We assume that $r_c\geq 2r_s$, i.e., if two robots can see the same target, then they can also communicate. 
The robots need to select their actions based on only local information. 

\begin{figure}[t]
\vspace{2 mm}
\centering{
{\includegraphics[width=0.6\columnwidth,trim= 0cm .0cm 0 0cm,clip]{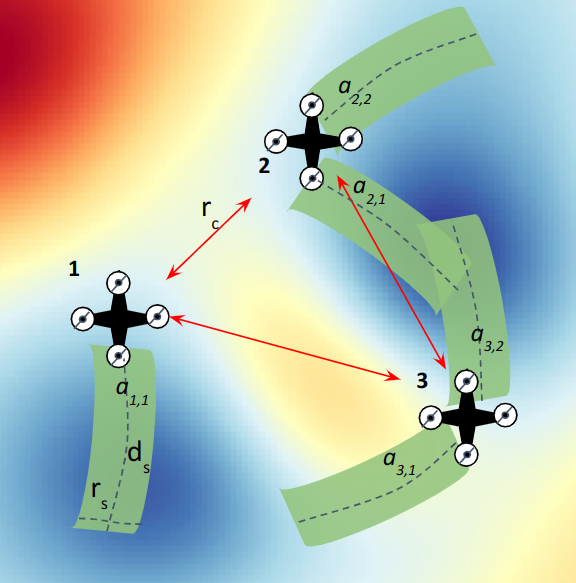}}
\caption{\small An illustrative example: at a given time step, each robot $R_i$ must choose a motion primitive $a_{i,k}$ (dashed curves). The background map shows areas with high target density with blue and low target density with red. Here, $R_1$ has one motion primitive $\{a_{1,1}\}$, $R_2$ has two primitives $\{a_{2,1}, a_{2,2}\}$, and $R_3$ also has two motion primitives $\{a_{3,1}, a_{3,2}\}$. The size of the \textit{coverage map} depends on the robot's sensing range $r_s$ and moving distance $d_s$. As $R_2$ and $R_3$ have overlapping coverage maps, they must communicate with each other using communication link (red arrows) of range $r_c$, to choose actions that can maximize the total coverage. \label{fig:uav_trajectory}}
}
\end{figure}

Each robot has access to a local \textit{coverage map}, which gives the \emph{predicted} occupancy of targets near the robot (specifically, targets that can be covered by its motion primitives). Any overlap in covering the same set of targets results in the targets being counted as \textit{covered} only once. We show an example in Figure~\ref{fig:uav_trajectory} where robot 2 and robot 3 may end up tracking the same target.  Thus, a robot must collaborate with others to minimize overlap in motion for efficient coverage. To do so, the robot must also share its local coverage map with others. It is also important to share a compact representation of the map to reduce the bandwidth requirement of the algorithm. 

Our main contribution is \DTWOCOPLAN{}, which solves both problems simultaneously. It consists of an map encoder that comes up with a compact representation of each robot's coverage map, an information aggregator, followed by an action selector. Furthermore, since \DTWOCOPLAN{} is differentiable, we can combine it with a Differentiable Map Predictor (\DMP{}), that takes as input the history of observations from a robot and predicts the coverage map of where the targets are going to be when the robots move.

\section{Methods}\label{sec:methods}
We present a differentiable, decentralized coverage planner called \DTWOCOPLAN{} to efficiently solve the multi-robot coverage problem by predicting the best action for a robot given its local coverage map. It can be integrated with any differentiable map predictor (\DMP{}), to solve tasks where direct observations are not available. We design \DTWOCOPLAN{} as a combination of three sub-modules:

\subsection{Map Encoder} This module takes the robot's coverage map as input and transforms it into a feature vector that can be share with the robot's neighbors. We implement this module using a multi-layer Convolutional Neural Network (CNN), consisting of convolution, pooling, and ReLU activation layers. The input to the encoder is the coverage map as a single channel image of size $G \times G$. The output features from the CNN are flattened into a vector of size $H\times 1$ before sharing with the neighbors. This also allows for compressing the local maps making it efficient to communicate them to other robots. We choose CNN as the encoder here over a fully-connected neural network as it allows for a richer representation than the pre-processed inputs required for the latter as used in prior work~\cite{zhou2021graph}. Furthermore, we do not need to limit the maximum number of targets as input in our representation.

\subsection{Distributed Feature Generator} This part of the network enables sharing of the map encoding features with a GNN. GNN enables feature aggregation for each graph node through neural networks, allowing distributed execution. The information can be shared with $K$-hop communication to the neighbors identified using the adjacency matrix. The output of this module summarize the information from the neighbors as a vector, enabling informed decision-making in the next step. 

\subsection{Local Action Selector} The last module of \DTWOCOPLAN{} is responsible for prescribing the best action to the robot based on the information gathered from the neighbors in the previous step. We implement this module as a Multi-Layer Perception (MLP) which outputs a $|A|$-dimensional output, denoting the fitness of each action, $a_i\in A$. During the training the loss is calculated as cross-entropy over these outputs with the ground truth actions. Thus, this module enables the gradient flow for end-to-end training for \DTWOCOPLAN{}.

For training \DTWOCOPLAN{}, we use a centralized greedy algorithm as the expert algorithm to generate the target actions. The centralized greedy algorithm has access to global information (i.e., the global coverage map) and can therefore make much more informed decisions. In fact, it is known that the centralized greedy algorithm is within $\sim 66\%$ of the centralized optimal which eliminates the need to run optimal, brute force search algorithm that is infeasible for generating training data for large number of robots. The expert algorithm evaluates the coverage by each robot-action pair and selects the pair with the highest value. The selected robot and the covered targets are removed from consideration and the process is repeated till each robot is assigned an action. The algorithm has a time complexity of $\mathcal{O}(n^2)$ for $n$ number of robots. We refer to this algorithm as \EXPERT{}.

\begin{figure*}[t]
\centering{
{\includegraphics[width=0.95\linewidth,trim= 0cm .0cm 0 0cm,clip]{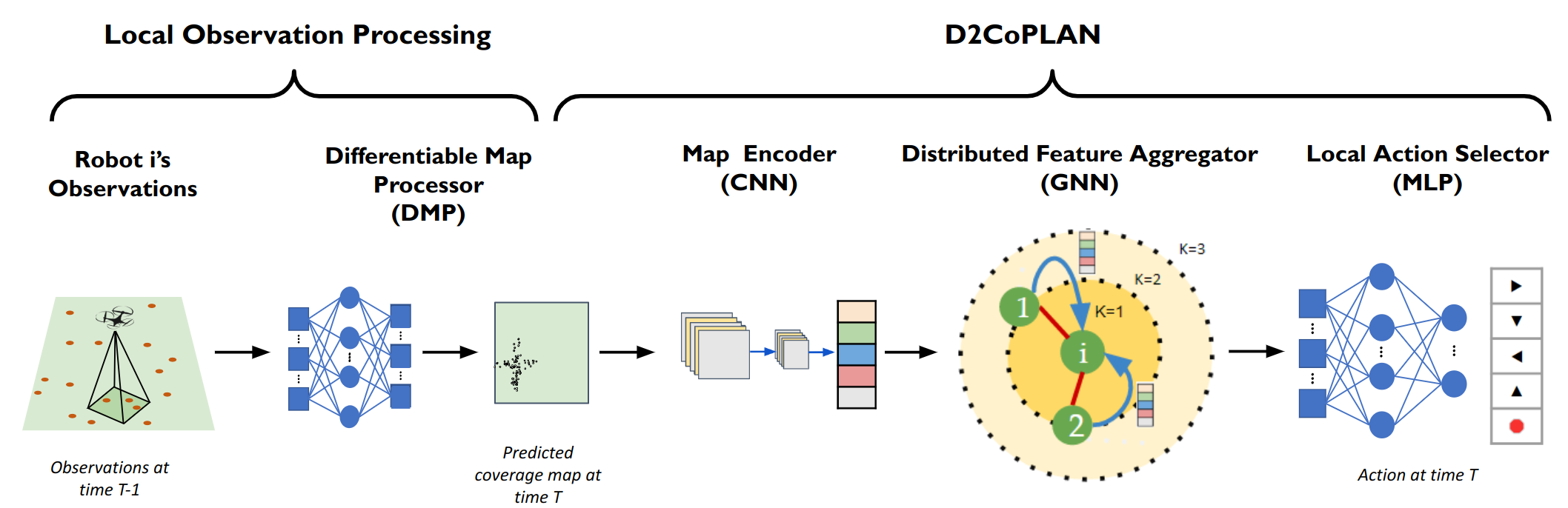}}
\caption{\small Overview of our approach from a robot's perspective: first the local observations are processed to generate the current coverage map. This can be done with the Differentiable Map Processor (\DMP{}). \DTWOCOPLAN{} takes the coverage map as the input and processes it to first generate compact feature representation, with \textit{Map Encoder}; shares the features with its neighbors, using the \textit{Distributed Feature Aggregator}; and then selects an action using the aggregated information, with the \textit{Local Action Selector}. The abbreviations in the parentheses for \DTWOCOPLAN{}'s sub-modules indicate the type of neural network used in their implementation. \label{fig:overview}}
}
\end{figure*}

\subsection{Differentiable Map Predictor}
To transform the robot's observation to coverage maps, we introduce a map predictor module. To allow integration with \DTWOCOPLAN{} in order to learn the transformation we use a differentiable map predictor (\DMP{}). The design of \DMP{} depends on the task at hand, and can be realized with neural networks. For example, if the task is defined as maximizing coverage with moving targets, \DMP{} can be implemented as a recurrent neural network. We use CNN to solve this task by stacking the historical observations as a multidimensional image and train it with a pre-trained \DTWOCOPLAN{} over the expert actions. This module is optional and we can use the ground truth coverage map for action selection, if available.

\section{Experiments and Results}\label{sec:experiment_results}
\subsection{Experiment Setup}\label{sec:setup}
In our experiments, We use \DTWOCOPLAN{} trained over $N=20$ robots. To generate the training data, we use a grid with $G=100$ i.e., a grid with size $100 \times 100$. The target coverage maps are generated using a mixture of Gaussian to simulate low and high density areas. The intuition is to mimic real-life situations such as animals density being higher closer to water holes and lower around ditches in a forest. For this, we choose a random number of Gaussian components in the range $[10,30]$ with the standard deviation for each uniformly sampled from the set $\{20, 30, 40, 50\}$. The locations of the means are selected uniformly at random on the grid. Some of the components are randomly inverted by multiplying by $-1$ to simulate lower density regions. The probability density obtained by summing up the components is then normalized to obtain a categorical probability density function over the grid. As the last step, we sample locations using this density function to fill $15\%$ of the grid cells to represent the target locations. We simulate linear motion for the targets with randomly chosen initial velocity.

The robot locations are randomly selected on the grid. The action set for each robot consists of 5 actions, one per cardinal direction and one to stay in place. The sensing range $r_s=6$, each action moves a distance of $d_s=20$, and communication range is $r_c=20$. With our choice of action primitives, the coverage map looks like a rectangular field on the grid of size $G \times G$, where only the target within the coverage map are visible. The communication is limited to 1-hop only. We generate total 40000 maps and run  \EXPERT{} on each to obtain the target actions. From this dataset, $60\%$ instances are used for training, $20\%$ are used for validation and the rest are used for testing. 

\textit{Map encoder} is implemented as a 3-layer CNN (Conv$\rightarrow{}$ReLU$\rightarrow{}$Maxpool) with intermediate output features of size $4$, $8$ and $16$. The final output is flattened to a vector of size $1600$. This vector acts as a compresses map representation. For \textit{Distributed feature aggregator}, we use implementation by Li et al.~\cite{li2020graph} with 2 graph layers of $512$ and $128$ nodes and ReLU activation. \textit{Local action selector} is implemented as a single layer fully connected network, directly predicting the actions. We also use dropout of $20\%$ in the CNN layers and after the GNN to regularize the network. We train the network on an Nvidia GeForce RTX 2080Ti GPU with 11GB of memory for 1500 epochs and use the network weights with the minimum validation loss for evaluation. 

\subsection{Evaluation}\label{sec:evaluation}
An efficient distributed planner must have some desirable properties: it should run faster than the centralized algorithm, while achieving coverage within a reasonable margin of the centralized algorithm; and it should scale well with varying number of agents by generalizing beyond the settings it is trained on. In this section, we present empirical evidence that \DTWOCOPLAN{} has the aforementioned desirable properties. We go one step further and show that \DTWOCOPLAN{} scales better than even \DG{}. Finally, we demonstrate the advantages of a \textit{differentiable} design. Specifically, we show that \DTWOCOPLAN{} performs better when combined with \DMP{} than \DG{}. 

\subsubsection{Comparisons with \EXPERT{}} We begin by comparing the coverage performance (number of targets covered) and runtime of \DTWOCOPLAN{} with the \EXPERT{} which is the centralized greedy algorithm that \DTWOCOPLAN{} is trained using. In this set of experiments, we use the ground truth coverage map as inputs since our focus is on evaluating the planner. In subsequent experiments, we will evaluate the effect of the map predictor on the coverage task.

\DTWOCOPLAN{} was trained on a dataset of 20 robots in a grid of size $100 \times 100$. We compare the two algorithms with increasing number of robots (from $4$ to $50$) in the same grid. We run 1000 Monte-Carlo simulations for each setting. 

% In this comparison, Our key focus here is on the runtime as the \textit{Expert} algorithm's main disadvantage is that its runtime increase quadratically with the increase in the number of robots. is severely affected with increasing number of agents. While \textit{Expert} is expected to achieve a higher covearge than \textsc{D2CoPLAN} because of the difference in their scope of information, we compare the coverage as well to highlight that \textsc{D2CoPLAN} performs reasonably well. 

\begin{figure*}[t]
\centering
\subfigure[Scaling in terms of Coverage]
{\includegraphics[width=0.9\columnwidth]{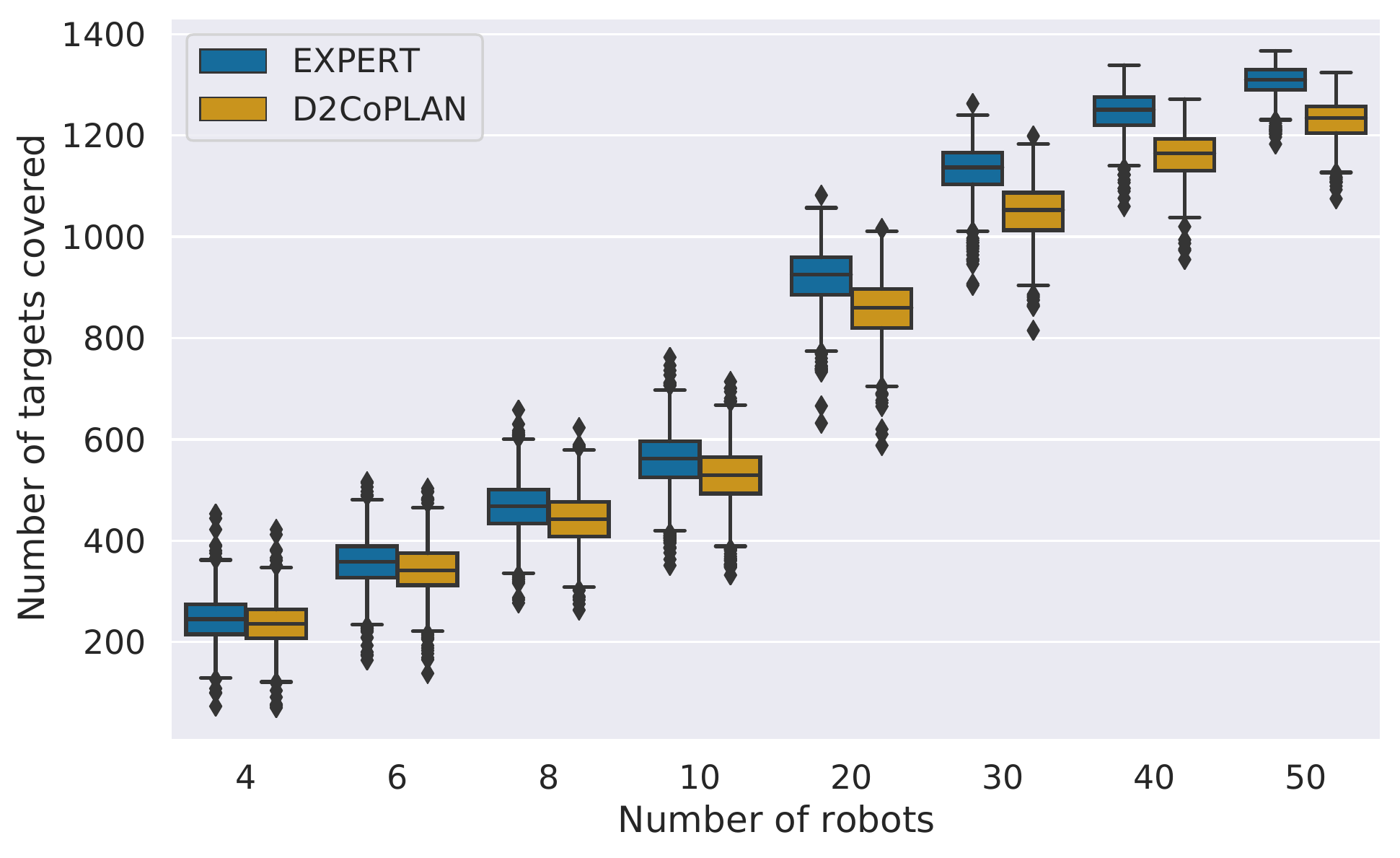}}~~~
\subfigure[Scaling in terms of time]
{\includegraphics[width=0.9\columnwidth]{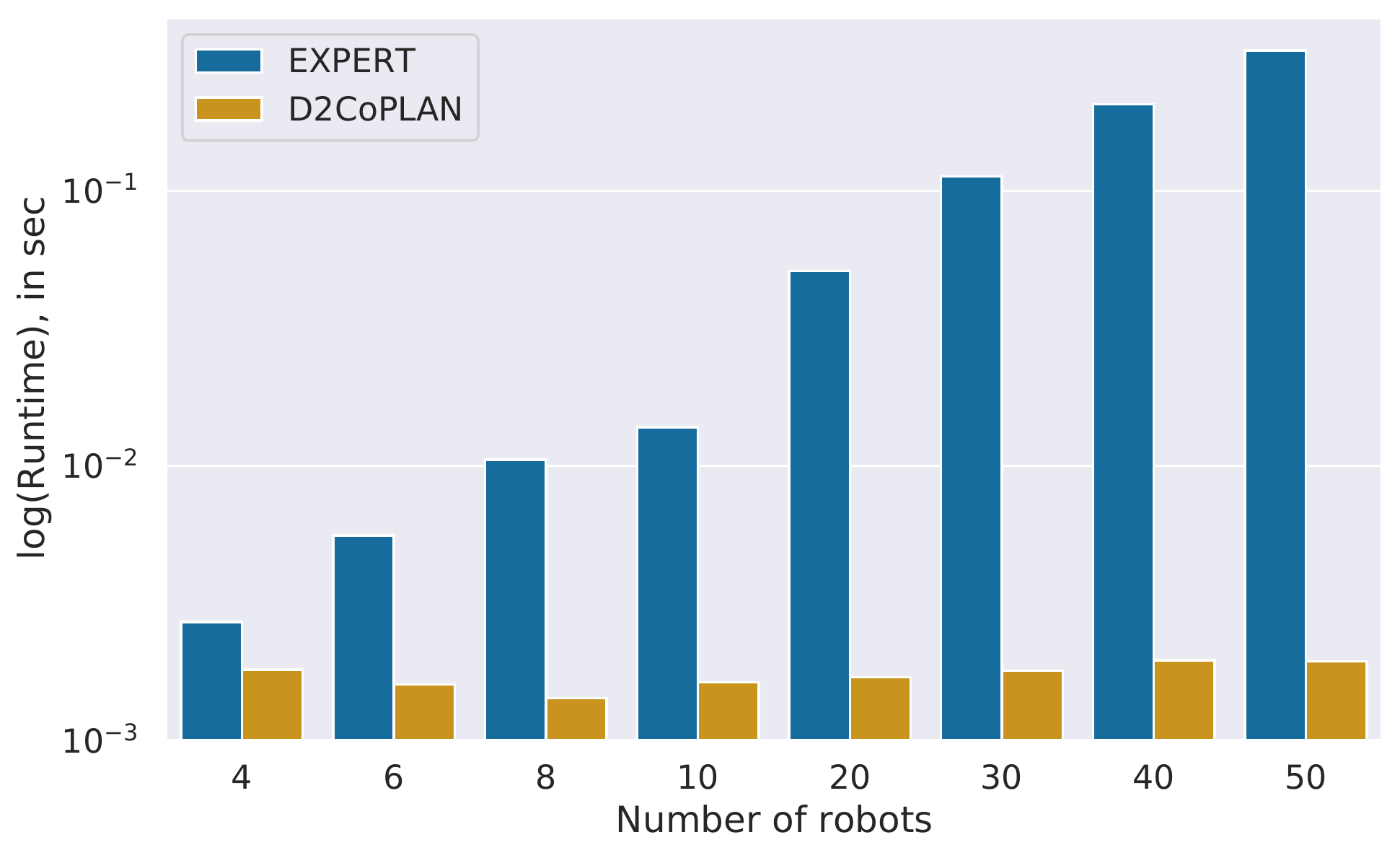}}
\caption{\small Comparison of \EXPERT{}, \DTWOCOPLAN{}, and \texttt{Random} in terms of running time (plotted in $\log$ scale) and the number of targets covered, averaged across 1000 Monte Carlo trials. \DTWOCOPLAN{} was trained on 20 robots. \DTWOCOPLAN{} is able to cover ~92\%-96\% of the targets covered by \EXPERT{}, while running at a much faster rate.}
\label{fig:expert_vs_gnn}
\end{figure*}

The results for this evaluation are shown in Figure~\ref{fig:expert_vs_gnn}. \DTWOCOPLAN{} has a clear advantage in terms of runtime and the advantage increases as the number of robots increases. For example, with 50 robots, \DTWOCOPLAN{} is more than two orders of magnitude faster than \EXPERT{}. This is not surprising since \EXPERT{} is a centralized algorithm whose runtime scales quadratically with the number of robots whereas \DTWOCOPLAN{} is a decentralized algorithm. In addition to being significantly faster, we also observe that \DTWOCOPLAN{} covers $~92\%$ of the targets as the \EXPERT{}, despite each robot having only a limited amount of information.

\subsubsection{Comparisons with \DG{}}
\begin{figure*}[t]
\centering
\subfigure[Scaling in terms of Coverage]
{\includegraphics[width=0.9\columnwidth]{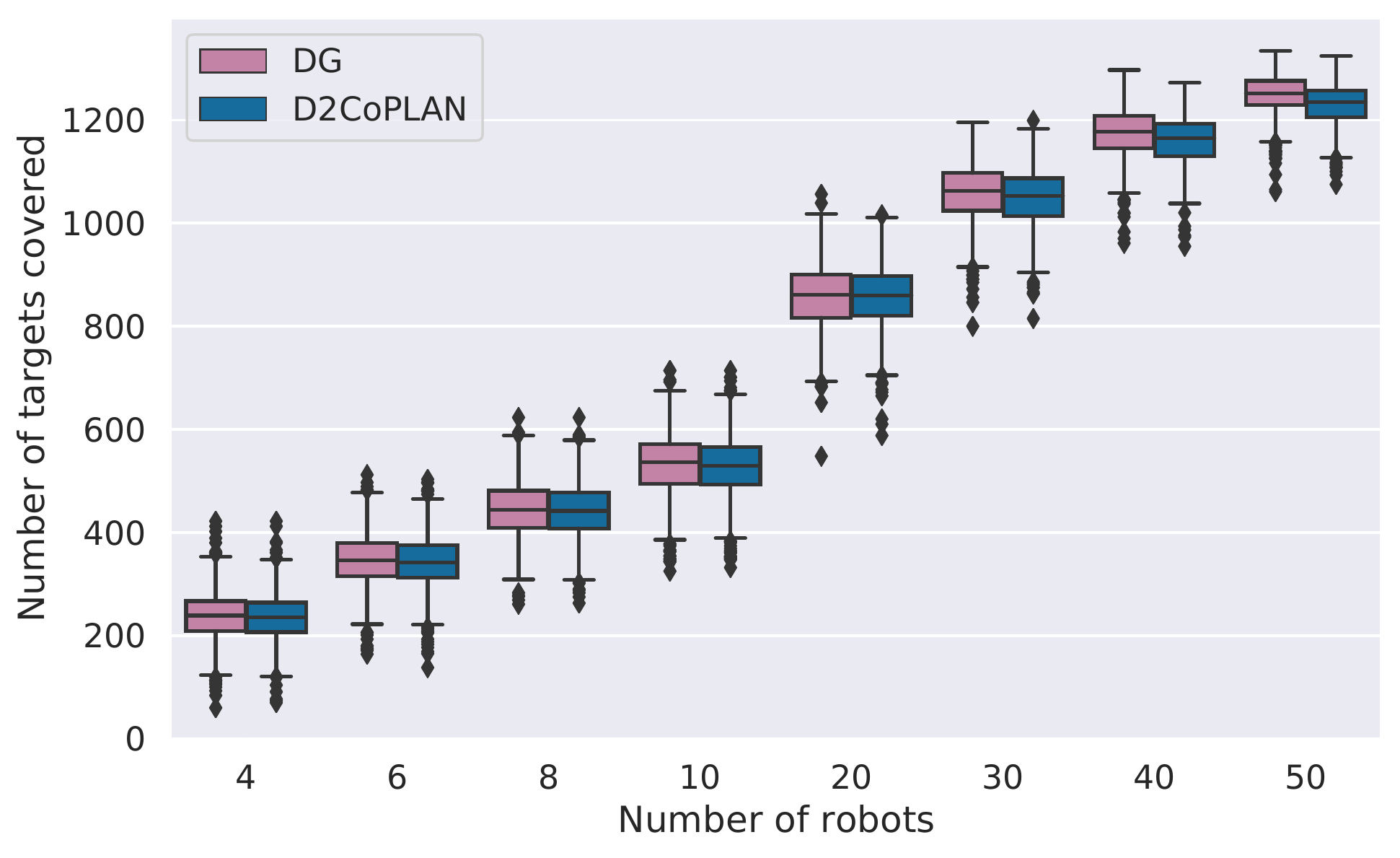}}~~~
\subfigure[Scaling in terms of time]
{\includegraphics[width=0.9\columnwidth]{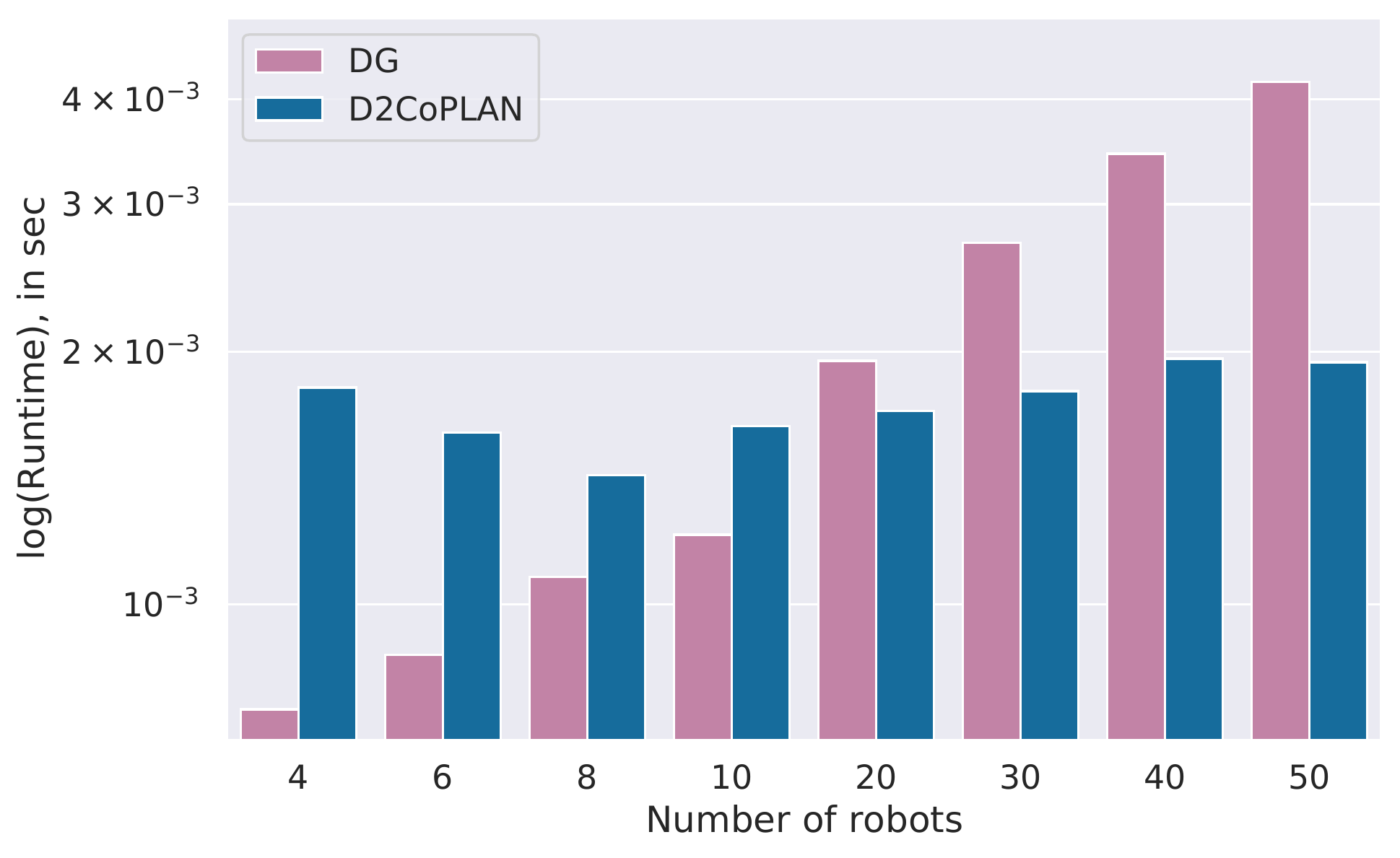}}
\caption{\small Comparison of \DTWOCOPLAN{}, and \DG{} in terms of running time (plotted in $\log$ scale) and the number of targets covered, averaged across 1000 Monte Carlo trials. \DTWOCOPLAN{} was trained on 20 robots. \DTWOCOPLAN{} is able to cover almost same number of targets as \DG{}. \DG{} is faster for fewer number of robots, but as the number of robots increase, \DTWOCOPLAN{} scales better than it.}
\label{fig:dg_vs_gnn}
\end{figure*}

Next, we compare \DTWOCOPLAN{} with a classical decentralized algorithm, \DG{}. In \DG{}, each robot  chooses its own action by running a greedy algorithm but only on the set that includes itself and its immediate neighbors (hence, decentralization). As shown in Figure~\ref{fig:dg_vs_gnn}, \DTWOCOPLAN{} and \DG{} perform almost the same in terms of the number of targets tracked. However, the real advantage of \DTWOCOPLAN{} comes in the runtime where we observe it becomes much faster than \DG{} as the number of robots increase (e.g., with 50 robots, \DTWOCOPLAN{} is almost twice as fast). While both algorithms are decentralized, \DG{} still requires running a greedy algorithm over the local neighborhood of each robot which increases the runtime as the density of the robots increase. Furthermore, in Section~\ref{sec:dmp_d2coplan} we show that \DTWOCOPLAN{} outperforms \DG{} even in terms of coverage performance when the true coverage map is not given. 

\subsubsection{Generalization}
Next, we evaluate the generalization capability of \DTWOCOPLAN{} beyond the scenario it has been trained on. We test two types of generalization: (1) across number of robots; and (2) across density of the targets (i.e., coverage map) in the environment. For both tests, we train on a specific number of robots (1) or target density (2) and test with a different number of robots (1) or target density (2). We summarize these results in Table~\ref{tab:general_robots} and Table~\ref{tab:general_density} obtained over 1000 Monte-Carlo runs.

% For (1), we train two variations of $D2CoPLAN$ for datasets built using 10 robots and 30 robots and check the percentage of targets covered by them with respect to \textit{Expert}. For (2), we report this metric for target density of $5\%$, $15\%$, $25\%$ and $50\%$. \textsc{D2CoPLAN} was trained on a map with $15\%$ target density.

We observe that \DTWOCOPLAN{} generalizes well in both cases. Table~\ref{tab:general_robots} shows the coverage performance when trained on the number of robots given in the row and tested on the number of robots given in the column. We see that in most cases, the performance remains unchanged. The network trained on $10$ robots sees a slight drop in performance on other test configurations but still covers around $~90\%$ of the targets covered by \EXPERT{}. 

\DTWOCOPLAN{} also generalizes well across varying target density as shown in Table~\ref{tab:general_density}. We observe that \DTWOCOPLAN{} trained with a target density of $15\%$ performs almost the same when tested on other target densities. The performance is $\sim 93\%$ of the \EXPERT{} in all cases but $5\%$ density (where it is $\sim 91\%$), which we believe is caused by fewer number of available targets, increasing the gap in the performance of the compared algorithms. These results validate the claim that \DTWOCOPLAN{} trained under one type of scenario generalizes to other deployment scenarios. 

\begin{table}
\vspace{1mm}
\centering
{\renewcommand{\arraystretch}{1.5}
\begin{tabular}{|l|*{3}{c|}}\hline
\backslashbox{Train}{Test}
&\makebox[2em]{10 Robots}&\makebox[2em]{20 Robots}&\makebox[3em]{30 Robots}\\\hline
\makebox[6em]{10 Robots} &93.95\% &91.12\% &89.71\% \\\hline
\makebox[6em]{20 Robots} &94.38\% &93.17\% &92.60\% \\\hline
\makebox[6em]{30 Robots} &93.25\% &93.47\% &93.73\% \\\hline
\end{tabular}}
\vspace{2mm}
\caption{Percentage of the targets covered (the average across 1000 trials) with respect to \EXPERT{} by \DTWOCOPLAN{} trained and tested with varying numbers of robots.}
\label{tab:general_robots}
\vspace{-2mm}
\end{table}

\begin{table}
\vspace{1.5 mm}
\centering
\begin{tabular}{ | c | c|} 
\hline
{Target Density} & {Relative coverage}\\
\hline
$5\%$ & 91.40\%\\
\hline
$15\%$ & 93.20\%\\
\hline
$25\%$ & 93.46\%\\
\hline
$50\%$ & 93.46\%\\
\hline
\end{tabular}
\caption{Percentage of the targets covered (the average across 1000 trials) with respect to \EXPERT{} by \DTWOCOPLAN{} across varying target density maps.}
\label{tab:general_density}
\end{table}

\subsubsection{Prediction and Planning}\label{sec:dmp_d2coplan}
A key advantage of \DTWOCOPLAN{} is its differentiablity, allowing \DTWOCOPLAN{} to be combined with other gradient-based learning methods to solve challenging problems in an end-to-end manner. In this section, we evaluate how the differentiable map predictor can be trained along with the differentiable planner (\DTWOCOPLAN{}) and compare it with \DG{}.

% Consider the problem of moving targets: in a multi-robot coverage problem, the agents observe the targets in their field-of-view at discrete time intervals $0$ to $T-1$ and must move at timestep $T$ to capture the targets, while maximizing the joint coverage. Unlike the previous setting, the robots must anticipate the next location of the targets and select the actions. 

So far, we have used the ground truth coverage map as input to the planners. Now, we consider a scenario where the input consists of the observations of the targets over the past timesteps. The true motion model of the robots is not known to the robots. Therefore, they need a predictor to estimate the positions of the targets over the planning horizon which can then be used by \DG{} or \DTWOCOPLAN{}. 

%  The robots observe the targets for 3 timesteps. The status of the targets at $T=4$ is unknown to the robots.  

% In order to solve this problem, we requires a \textit{mapper} and a \textit{planner}. 

Here, we use a \DMP{} to learn the motion model. The targets move with a linear velocity selected randomly at the start of the episode (unknown to the planner). To show the advantage of having a decentralized planner, we compare three methods: (1) an \ORACLE{} i.e., the ground truth map as the mapper along with \EXPERT{} as the planner; (2) \DMP{} as the learnable mapper with \DG{} as the planner; and (3) \DMP{} as the learnable mapper with \DTWOCOPLAN{} as the planner. In (2), \DMP{} is trained from scratch where in (3) \DMP{} is trained by backpropagating the loss from \DTWOCOPLAN{}. \DTWOCOPLAN{} itself is frozen and aids \DMP{} in learning better representations for action prediction. The three settings present different combinations classical and learning-based approaches. 

Coverage maps observed over last 3 time steps are used as input to \DMP{} and it predicts the map at the next time step. We use a 4-layer CNN with 8, 16, 4 and 2 channels as \DMP{}. We keep the map size same across each layer to avoid information loss and predict the occupancy probability of each cell as a two channel map. The probability map thus obtained is used as input to the planner.  We trained \DMP{} over 2000 epochs with 5 examples of 20 robots (i.e., 100 training instances) in each. Given that most of the cells in the coverage map will be zero, we weigh the cross-entropy loss by a ratio of 1:10 for free and occupied cells. The action prediction loss for (3) is the unweighted cross-entropy loss. 

\begin{figure}[t]
\centering
{\includegraphics[width=0.90\columnwidth]{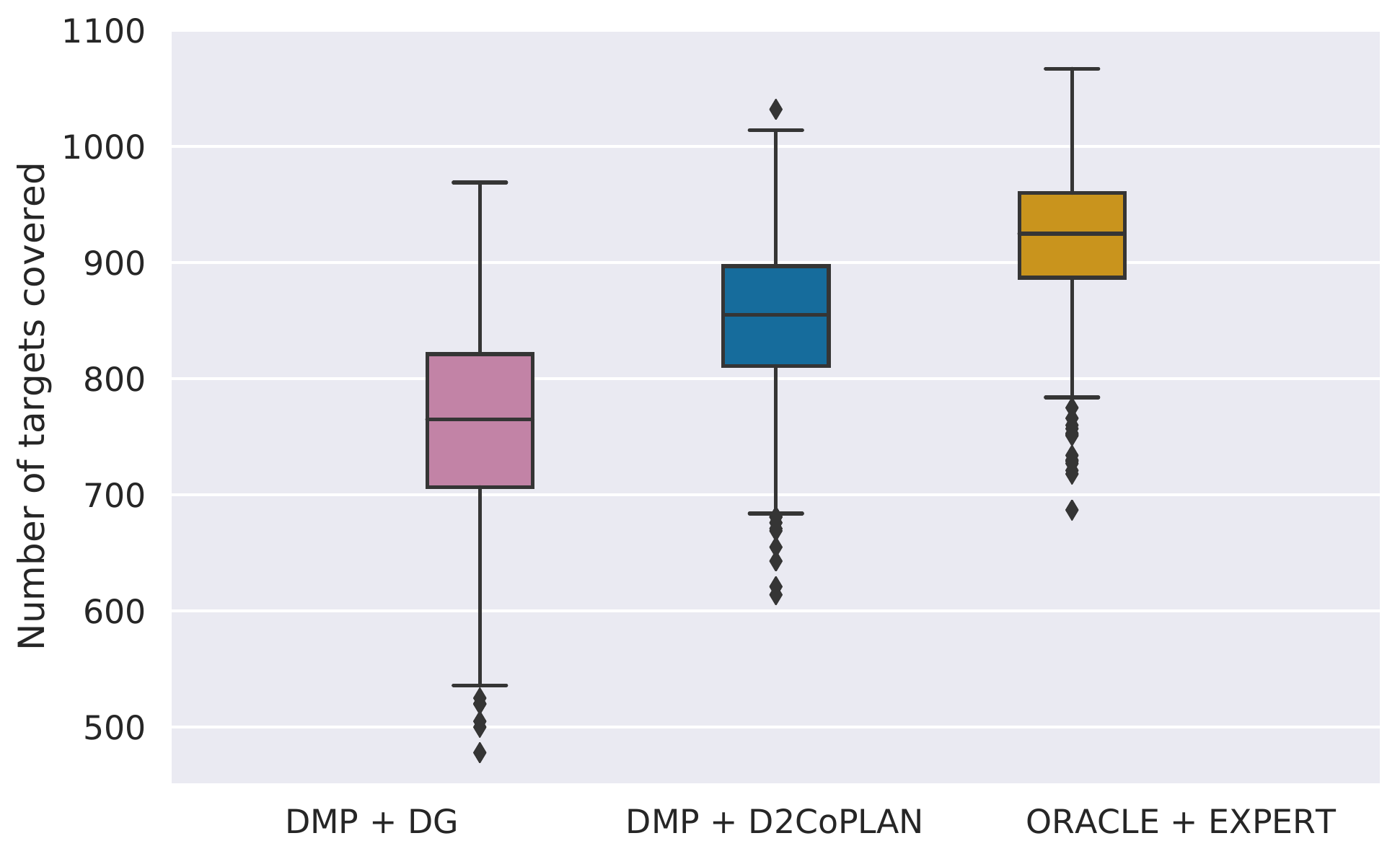}
}
\caption{\small Comparison of coverage highlighting the effect of using \DTWOCOPLAN{}, a differentiable planner to aid learning for a differentiable map predictor (\DMP{}), which works better than the \DMP{} trained standalone.}
\label{fig:downstream_comp}
\end{figure}

Figure~\ref{fig:downstream_comp} shows a comparison of the three approaches and provides evidence for benefit of using a differentiable planner to realize end-to-end learning. The combination of \DTWOCOPLAN{} and \DMP{} is better compared to \DG{} and \DMP{}, despite \DMP{} in the latter being trained on ground truth. We attribute this to the fact that \DTWOCOPLAN{} and \DMP{} form a differentiable chain which allows \DMP{} to be trained directly on the downstream task (action selection) rather than on just map prediction. \DG{} and \DMP{}, on the other hand, are not a differentiable chain and thus \DMP{} cannot be trained on the downstream task directly.

We further explore this by comparing 3 ways of training \DMP{} when used in conjunction with \DTWOCOPLAN{}: (1) \DMP{} and \DTWOCOPLAN{} are trained together from scratch; (2) \DMP{} and \DTWOCOPLAN{} are trained individually and then used together; and (3) \DTWOCOPLAN{} is first trained and then \DMP{} is trained on loss from \DTWOCOPLAN{} while \DTWOCOPLAN{} is frozen.

\begin{figure}
\centering
{\includegraphics[width=0.90\columnwidth]{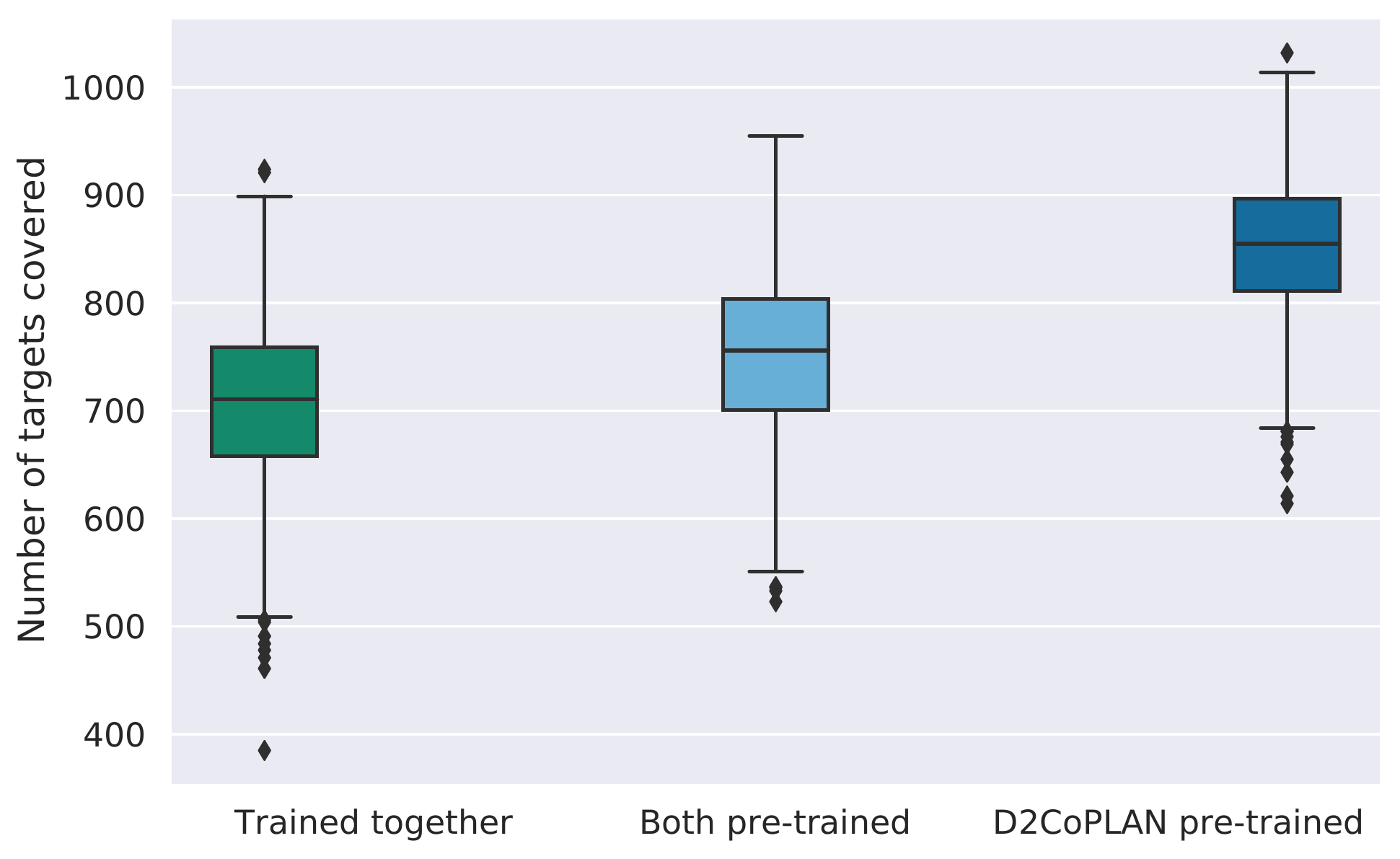}}
\caption{\small An ablation study for \DMP{} and \DTWOCOPLAN{}. The plot shows results for the scenarios where there parts and trained together or in isolation.}
\label{fig:downstream_ablation}
\end{figure}

Figure~\ref{fig:downstream_ablation} shows the comparison of all three methods. The third approach outperforms the other two. This demonstrates the advantage of having a differentiable planner. Using a pre-trained and frozen \DTWOCOPLAN{} and training directly on the downstream task loss, allows \DMP{} to learn patterns beneficial for action prediction and not just for map prediction. If both modules are trained from scratch in an end-to-end manner, they may need more time to learn the same behavior. The third approach also does not require ground truth motion models for the targets to be available for training \DMP{}. While in this paper we use the ground truth to generate the expert solutions used in training \DTWOCOPLAN{}, in general, one can use any other expert algorithm such as human inputs to train \DTWOCOPLAN{} which does not need ground truth target motion.

% Better performance of the \textit{Both pre-trained} scenario helps in delivering the key message here: \textit{we do not present GNN to solve the coverage problem; we present a planner based on GNN}. Having a differentiable planner helps with the downstream tasks.

\section{Conclusion}\label{ref:conclusion}
We presented \DTWOCOPLAN{}, a differentiable, decentralized target coverage planner for multi-robot teams. Our experimental results show that \DTWOCOPLAN{} is more scalable than the classical decentralized algorithm that is used for such tasks while performing closer to the centralized algorithm. Furthermore, due to the fact that it is a differentiable planner, we can combine this with other differentiable modules (e.g., a coverage map predictor) to yield better performance than the classic counterparts. These results present an encouraging path forward for multi-robot coordination tasks. Our immediate work is evaluating \DTWOCOPLAN{} for more complex tasks. In this paper, we train \DTWOCOPLAN{} in a supervised setting. We are also working on training \DTWOCOPLAN{} with reinforcement learning. Finally, an interesting avenue for extension is where we learn not just \emph{what} to communicate with other robots (as we do in this paper) but also \emph{who} to communicate with. 
% \section*{ACKNOWLEDGMENT}

% %%%%%%%%%%%%%%%%%%%%%%%%%%%%%%%%%%%%%%%%%%%%%%%%%%%%%%%%%%%%%%%%%%%%%%%%%%%%%%%%

% References are important to the reader; therefore, each citation must be complete and correct. If at all possible, references should be commonly available publications.

{\small
\bibliographystyle{ieee_fullname}
\bibliography{bibfile}
}

\end{document}